# Evaluating Language Tools for Fifteen EU-official Under-resourced Languages

**Diego Alves, Gaurish Thakkar, Marko Tadić**
University of Zagreb, Faculty of Humanities and Social Sciences
Ivana Lučića 3, 10000 Zagreb, Croatia
dfvalio@ffzg.hr, gthakkar@m.ffzg.hr, marko.tadic@ffzg.hr

**Abstract**
This article presents the results of the evaluation campaign of language tools available for fifteen EU-official under-resourced languages. The evaluation was conducted within the MSC ITN CLEOPATRA action that aims at building the cross-lingual event-centric knowledge processing on top of the application of linguistic processing chains (LPCs) for at least 24 EU-official languages. In this campaign, we concentrated on three existing NLP platforms (Stanford CoreNLP, NLP Cube, UDPipe) that all provide models for under-resourced languages and in this first run we covered 15 under-resourced languages for which the models were available. We present the design of the evaluation campaign and present the results as well as discuss them. We considered the difference between reported and our tested results within a single percentage point as being within the limits of acceptable tolerance and thus consider this result as reproducible. However, for a number of languages, the results are below what was reported in the literature, and in some cases, our testing results are even better than the ones reported previously. Particularly problematic was the evaluation of NERC systems. One of the reasons is the absence of universally or cross-lingually applicable named entities classification scheme that would serve the NERC task in different languages analogous to the Universal Dependency scheme in parsing task. To build such a scheme has become one of our the future research directions.

## 1. Introduction

The goal of the Marie Skłodowska-Curie Innovative Training Network CLEOPATRA (Cross-lingual Event-centric Open Analytics Research Academy)[1] is to offer a unique interdisciplinary and cross-sectoral research and training programme, which will explore how we can begin to analyse and understand major events and their relations presented in digital media. It will facilitate advanced cross-lingual processing of textual and visual information related to key contemporary events at scale and will develop innovative methods for efficient and intuitive user access and interaction with multilingual information. This ambitious goal heavily relies on the necessary prerequisite: the establishment of online available language processing chains (LPCs) for at least 24 EU-official languages of which many are also under-resourced. The components of LPCs all belong to the BLARK-modules (Krauwer, 2003) and cover tokenisation, sentence splitting, PoS/MSD-tagging, lemmatisation, named entity recognition and classification (NERC) and dependency parsing. On the results of LPCs regularly applied to the sets of large publicly available text streams (news, blogs, posts, etc.), the CLEOPATRA knowledge processing pipeline will build upon and provide additional value such as knowledge graphs, entity linking with LOD, RDF triples, other Semantic Web processing, etc. We expect that one of the final results of CLEOPATRA relevant to the NLP community will be the enhancement of the availability and performance of human language technologies for under-resourced languages.

In order to get the clear(er) picture of currently available tools and training/testing resources for under-resourced languages, this paper concentrated on under-resourced languages only and aims to find and test these tools to verify their performance.

The paper is composed as follows: Section 2 discusses the selection of language tools and languages that were used in this evaluation; Section 3 describes the design of the evaluation campaign; Section 4 lists the processing tasks and their individual results; in Section 5 we discuss some results and in Section 6 we provide conclusions and possible future directions for research.

## 2. Tools and Languages Selection

The main objective of this study is to produce an overview of the current state of available language tools and respective needed training and testing resources for under-resourced EU-official languages, while also checking whether the results in the real-life comply with reported in the literature.

Our aim is to analyse the whole language processing chain, from raw text to CoNLL-U file and for that reason we have selected three platforms (or frameworks) with existing trained models available for most of the European languages, thus including the fifteen under-resourced languages in our focus. These are listed in Table 1.

| Platform | Version | License |
|---|---|---|
| Stanford CoreNLP | 3.9.2 | GNU General Public License v3 |
| NLP Cube | - | Apache v2 |
| UDPipe | 1.2.0 | Mozilla Public License 2.0 |

Table 1: Selected Natural Language Processing Platforms.

---

[1] http://cleopatra-project.eu

In this paper, our focus is on under-resourced EU-official languages and the selection of languages has been made according to the classification presented by the META-NET Language Whitepaper series (Rehm et al. 2012). The chosen languages are classified by META-NET as either "Fragmentary" or "Weak/No" support. The other criteria for language selection were the existence of quality datasets (Universal Dependencies framework version 2.4 released on May 15, 2019)[2] and the availability of trained models for the three selected platforms.

Therefore, from the twenty-four official European Union languages, for this study we have chosen the following fifteen: Croatian, Czech, Danish, Estonian, Finnish, Greek, Hungarian, Irish, Latvian, Polish, Portuguese, Romanian, Slovak, Slovene, Swedish. Maltese, although being heavily under-resourced EU-official language, could not be included in this evaluation since trained models do not exist for all three selected platforms.

As NERC is also a crucial part of LPCs, for each of the fifteen languages listed above, we also conducted research on existing tools and data to complete our LPC overview. Due to the lack of resources, we have excluded Irish, Latvian, and Slovak from this part of our study.

## 3. Campaign Design

Unlike the shared task campaigns where different tools are being tested on the same data from different, often competing, centres, here we execute a post-hoc evaluation of what has been already presented. Our objective is to check whether published results are reproducible and to compare platforms' performance in terms of NLP metrics and processing speed.

For each language, we selected the available trained models downloadable from the official website of each platform. All of them used training data from the Universal Dependencies framework. UDPipe and StanfordNLP detail in which the exact UD corpus has been used and the respective UD version (v2.4 for UDPipe and v2 for StanfordNLP). NLP-Cube website does not provide precise information on the exact corpus used for the available trained models, although it confirms that models have been generated using UD v2.2 datasets. Boroş et al. (2018) present more detailed information about trained models using UD datasets, however, results in many cases are different from those displayed on the official website.

Each model, trained with a specific UD training corpus and downloaded from the official platforms' websites, was tested by processing the raw testing text coming from the same UD repository (v.2.4). The final processed file was directly compared to the available test standard provided by UD framework in terms of Precision, Recall and F1 measure for several NLP metrics.

As the UD datasets have no resources for NERC, we analysed the performance of the models on other publicly available datasets. Only the datasets which were not used for the training of the models were selected for testing.

## 4. Processing and Results

For each one of the chosen fifteen languages, we have processed the raw testing texts (UD 2.4) with the three mentioned platforms using the existing models available on their website by using the standard configuration proposed by each platform.

With these three platforms and starting from raw text, we have performed: sentence splitting, tokenisation, PoS/MSD-tagging, lemmatisation, and dependency parsing. The annotated files obtained were then evaluated and compared to the CoNLL-U standard test files from UD framework by using the python script proposed by the CoNLL 2018 Shared Task[3], which calculates the Precision, Recall and F1 Measure for the following metrics: Tokens, Sentences, Words, UPOS, XPOS, UFEATS, ALLTAGS, LEMMAS, UAS, LAS, CLAS, MLAS, and BLEX.

The tables presented in the next sections focus on the F1 Measure. The CoNLL-U standard test files were used exactly as they are presented in the UD framework, with their original tag sets and size (varying from one corpus to the others).

For each test, we have also measured the processing time in seconds and, subsequently, calculated processing speed by dividing the number of tokens and sentences of the test file by it (results in subsection 4.6).

All three platforms present in their official website evaluation metrics for the existing trained models of each one of the fifteen languages. Nevertheless, no information on the standard deviation or variance is available. Therefore, in order to evaluate the reproducibility of the official published results, we have checked the delta between the obtained metrics and the values present in the platforms' websites. Reproducibility is attested if the delta value is lower than 1% point, the value chosen as a first approach to highlight the main differences between published and tested results. However, NLP-Cube official results on the platform website do not come only from raw text processing, for this platform PoS/MSD-tagging and dependency parsing was conducted over a pre-segmented text.

For NERC, no pre-processing was performed on the tested texts. Most of the time the datasets were available in CoNLL format. Since different datasets had different numbers of tags, therefore the number of tags has been reduced to 3 namely **LOCATION**, **PERSON**, **ORGANISATION**, which are the ones available in all datasets. The remaining tags were replaced with the tag **OTHER**.

The following methodology was adopted for each language:
- Shortlist candidate datasets and available NERC tools.

---

[2] https://universaldependencies.org/

[3] https://universaldependencies.org/conll18/evaluation.html

- Cross-check whether the dataset used for training of a particular NERC tool is available.
- If present, check if the corresponding test set is available. If the test set is unavailable the whole dataset is removed. We perform this step in order to prevent reporting of train accuracy rather than test accuracy, which might occur when we sample from the whole dataset
- If no test set is available, we sample a small corpus from another candidate corpus.
- The test set is passed through the NERC tool which performs prediction by tagging every token of the input with a named entity class tag.
- We employ the script proposed in (Luz et al., 2018), which computes Precision, Recall, F1 measure based on individual tokens. The same script is used for all tests.
- In no scenario, a new model was trained from scratch.

The tables with all the obtained results from this evaluation campaign couldn't be presented within the technical limits of this paper stylesheet, so we decided to present them fully in the digital form and hosted online[4]. They represent an integral part of this research and could be consulted for more detailed insight whenever needed.

In the next subsections, we display in tables the cases for which we have observed a discrepancy between our data and the officially reported values.

In the following subsections, the tables present for each language first the result reported by the platform, followed by the results obtained in our tests. The cell/column names indicate whether the result is coming from the tool publication ("Source") or our experiment ("Test").

For NLP-Cube, as the official results are not coming from complete raw text processing and as there is no information about the origin of the testing data, our values are not comparable to theirs in our reproducibility analysis. The values we obtained for POS/MSD-tagging and dependency parsing for NLP-Cube are in all cases inferior to the announced results, which can be explained by our way of testing (from the raw text). NLP-Cube obtained results are also considerably lower than the ones from the other two platforms.

### 4.1 Tokenisation

For the tokenisation task, all results were coherent with the published ones from StanfordNLP and UDPipe, i.e. all results of our testing were within 1% point from the reported values. In the NLP Cube case, tokenisation is reproducible only for some test sets, which may indicate the provenience of the training data.

In Table 2, we present the comparative results for the tokenisation task between what has been published in the official NLP Cube platform website and what we obtained in our tests. The reported results correspond to the lines identified as "NLP-Cube Test Corpus".

| Language | Model | Test corpus | TOKENS |
|---|---|---|---|
| Croatian | hr-1.1 | NLP-Cube Test Corpus | 99.95 |
| | | hr_set-ud-test | 99.86 |
| Czech | cs-1.1 | NLP-Cube Test Corpus | 99.99 |
| | | cs_cac-ud-test | 99.97 |
| | | cs_cltt-ud-test | 83.39 |
| | | cs_fictree-ud-test | 99.98 |
| | | cs_pdt-ud-test | 99.47 |
| | | cs_pud-ud-test | 99.99 |
| Danish | da-1.1 | NLP-Cube Test Corpus | 99.82 |
| | | da_ddt-ud-test | 99.78 |
| Estonian | et-1.1 | NLP-Cube Test Corpus | 99.91 |
| | | et_edt-ud-test | 99.85 |
| | | et_ewt-ud-test | 98.5 |
| Finnish | fi-1.1 | NLP-Cube Test Corpus | 99.65 |
| | | fi_ftb-ud-test | 99.97 |
| | | fi_tdt-ud-test | 99.6 |
| | | fi_pud-ud-test | 99.61 |
| Greek | el-1.1 | NLP-Cube Test Corpus | 99.88 |
| | | el_gdt-ud-test | 99.91 |
| Hungarian | hu-1.1 | NLP-Cube Test Corpus | 99.88 |
| | | hu_szeged-ud-test | 98.41 |
| Irish | ga-1.0 | NLP-Cube Test Corpus | 99.56 |
| | | ga_idt-ud-test | - |
| Latvian | lv-1.0 | NLP-Cube Test Corpus | 99.66 |
| | | lv_lvtb-ud-test | 99.73 |
| Polish | pl-1.1 | NLP-Cube Test Corpus | - |
| | | pl_pdb-ud-test | 98.01 |
| | | pl_pud-ud-test | 97.89 |
| Portuguese | pt-1.1 | NLP-Cube Test Corpus | 99.75 |
| | | pt_bosque-ud-test | 99.79 |
| Romanian | ro-1.1 | NLP-Cube Test Corpus | 99.71 |
| | | ro_rrt-ud-test | 99.66 |
| Slovak | sk-1.1 | NLP-Cube Test Corpus | 99.95 |
| | | sk_snk-ud-test | 99.95 |
| Slovene | sl-1.1 | NLP-Cube Test Corpus | 99.87 |
| | | sl_ssj-ud-test | 99.95 |
| Swedish | sv-1.1 | NLP-Cube Test Corpus | 99.36 |
| | | sv_lines-ud-test | 99.97 |
| | | sv_pud-ud-test | 98.34 |
| | | sv_talbanken-ud-test | 99.36 |

Table 2: Tokenisation results for the NLP Cube platform. NLP-Cube Test Corpus indicates the reported values from the platform website. The other results come from our tests.

---
[4] https://tinyurl.com/wbp9wfq

### 4.2 Lemmatisation

Table 3 shows the results of the lemmatisation task for which the difference between the published results differs from the values obtained in our tests is larger than 1% point.

| Language | Source | Model | Test corpus | LEMMAS |
|---|---|---|---|---|
| Croatian | StanfordNLP | hr_set | hr_set-ud-test | 95.40 |
| Croatian | Test | hr_set | hr_set-ud-test | 97.21 |
| Finnish | UDPipe | finnish-ftb | fi_ftb-ud-test | 91.80 |
| Finnish | Test | finnish-ftb | fi_ftb-ud-test | 88.52 |
| Finnish | UDPipe | finnish-tdt | fi_tdt-ud-test | 96.60 |
| Finnish | Test | finnish-tdt | fi_tdt-ud-test | 86.88 |
| Greek | StanfordNLP | el_gdt | el_gdt-ud-test | 95.60 |
| Greek | Test | el_gdt | el_gdt-ud-test | 91.96 |
| Latvian | StanfordNLP | lv_lvtb | lv_lvtb-ud-test | 92.20 |
| Latvian | Test | lv_lvtb | lv_lvtb-ud-test | 89.03 |
| Polish | StanfordNLP | pl_sz | pl_pdb-ud-test | 95.94 |
| Polish | Test | pl_sz | pl_pdb-ud-test | 94.64 |
| Portuguese | StanfordNLP | pt_bosque | pt_bosque-ud-test | 96.88 |
| Portuguese | Test | pt_bosque | pt_bosque-ud-test | 85.66 |

Table 3: Lemmatisation results in cases where the discrepancy between published results and our tests is larger than 1% point. "Source" indicates whether the results come from the literature or our experiment.

### 4.3 POS/MSD-Tagging

For the evaluation of the reproducibility of the PoS/MSD-tagging, we present in Table 4. below the results for the ALLTAGS metric as described in the CoNLL 2018 Shared Task which takes into consideration UPOS, XPOS and UFEATS.

| Language | Source | Model | Test corpus | ALLTAGS |
|---|---|---|---|---|
| Croatian | StanfordNLP | hr_set | hr_set-ud-test | 91.14 |
| Croatian | Test | hr_set | hr_set-ud-test | 0.00 |
| Croatian | UDPipe | croatian-set-ud | hr_set-ud-test | 83.80 |
| Croatian | Test | croatian-set-ud | hr_set-ud-test | 90.10 |
| Finnish | StanfordNLP | fi_ftb | fi_ftb-ud-test | 94.27 |
| Finnish | Test | fi_ftb | fi_ftb-ud-test | 89.67 |
| Finnish | UDPipe | finnish-ftb-ud | fi_ftb-ud-test | 86.50 |
| Finnish | Test | finnish-ftb-ud | fi_ftb-ud-test | 88.93 |
| Finnish | UDPipe | finnish-tdt-ud | fi_tdt-ud-test | 95.70 |
| Finnish | Test | finnish-tdt-ud | fi_tdt-ud-test | 90.77 |
| Greek | StanfordNLP | el_gdt | el_gdt-ud-test | 94.22 |
| Greek | Test | el_gdt | el_gdt-ud-test | 86.43 |
| Latvian | StanfordNLP | lv_lvtb | lv_lvtb-ud-test | 86.20 |
| Latvian | Test | lv_lvtb | lv_lvtb-ud-test | 80.70 |
| Latvian | UDPipe | latvian-lvtb-ud | lv_lvtb-ud-test | 82.50 |
| Latvian | Test | latvian-lvtb-ud | lv_lvtb-ud-test | 83.60 |
| Polish | StanfordNLP | pl_sz | pl_pdb-ud-test | 93.77 |
| Polish | Test | pl_sz | pl_pdb-ud-test | 91.35 |
| Portuguese | StanfordNLP | pt_bosque | pt_bosque-ud-test | 94.34 |
| Portuguese | Test | pt_bosque | pt_bosque-ud-test | 81.87 |

Table 4: PoS/MSD-Tagging results in cases where the discrepancy between published results and our tests is larger than 1% point. "Source" indicates whether the results come from literature or our experiment.

### 4.4 Dependency Parsing

In terms of dependency parsing, we present in Table 5 the UAS and LAS obtained values compared to the ones from the official websites platforms when the difference for LAS is larger than 1% point.

| Language | Source | Model | Test corpus | UAS | LAS |
|---|---|---|---|---|---|
| Croatian | StanfordNLP | hr_set | hr_set-ud-test | 90.93 | 86.31 |
| Croatian | Test | hr_set | hr_set-ud-test | 92.94 | 89.37 |
| Czech | StanfordNLP | cs_fictree | cs_fictree-ud-test | 93.26 | 90.35 |
| Czech | Test | cs_fictree | cs_fictree-ud-test | 92.20 | 89.30 |
| Czech | StanfordNLP | cs_pdt | cs_pdt-ud-test | 91.85 | 89.60 |
| Czech | Test | cs_pdt | cs_pdt-ud-test | 90.39 | 88.10 |
| Finish | StanfordNLP | fi_ftb | fi_ftb-ud-test | 89.57 | 86.96 |
| Finish | Test | fi_ftb | fi_ftb-ud-test | 88.59 | 85.88 |
| Finish | UDPipe | finnish-tdt-ud | fi_tdt-ud-test | 85.30 | 80.00 |
| Finish | Test | finnish-tdt-ud | fi_tdt-ud-test | 80.49 | 76.85 |
| Greek | StanfordNLP | el_gdt | el_gdt-ud-test | 90.91 | 88.48 |
| Greek | Test | el_gdt | el_gdt-ud-test | 85.94 | 83.01 |
| Latvian | UDPipe | latvian-lvtb-ud | lv_lvtb-ud-test | 76.20 | 71.00 |
| Latvian | Test | latvian-lvtb-ud | lv_lvtb-ud-test | 79.30 | 74.31 |
| Polish | StanfordNLP | pl_sz | pl_pdb-ud-test | 96.00 | 94.09 |
| Polish | Test | pl_sz | pl_pdb-ud-test | 93.00 | 90.66 |
| Portuguese | StanfordNLP | pt_bosque | pt_bosque-ud-test | 90.25 | 87.98 |
| Portuguese | Test | pt_bosque | pt_bosque-ud-test | 73.40 | 68.83 |

Table 5: Dependency parsing results in cases where the discrepancy between published LAS results and our tests is larger than 1% point. "Source" indicates whether the results come from the literature or our experiment.

### 4.5 Processing Speed

In the following table, we present the mean processing speed for each platform and the associated standard deviation. The mean value was calculated by dividing the sum of the testing processing speeds by the number of conducted tests. The machine used for all tests has 8GB RAM and the Intel Pentium Quad-Core Processor N3710. All tests were conducted with the standard commands provided in the user's manual of each platform, no test was conducted by changing advanced testing parameters.

| | Mean Processing Speed (Tokens/s) | Standard Deviation | Mean Processing Speed (Sentences/s) | Standard Deviation |
|---|---|---|---|---|
| StanfordNLP | 53.2 | 24.6 | 3.5 | 1.7 |
| NLP-Cube | 5.6 | 0.7 | 0.3 | 0.1 |
| UDPipe | 381.1 | 45.2 | 24.6 | 9.8 |

Table 6: Mean processing speeds and standard deviation of each tested platform in terms of tokens and sentences per second.

### 4.6 Named Entity Recognition

Table 7 on the next page describes the available selected resources for the twelve languages that made through our shortlisting criteria.

| Language | Dataset | Entities | Size (Tokens) |
|---|---|---|---|
| Croatian | hr500k | P, L, O | 500k |
| Croatian | Vjesnik | P, L, O | 310k |
| Czech | Czech Named Entity Corpus 2.0 | 2-level Hierarchy of 46 tags | 8k sentences |
| Danish | DDT | P, L, O | 100k |
| Danish | WikiANN | P, L, O | 832k |
| Estonian | NER-tagger corpus | P, L, O, Facility, Product, Other | 217k |
| Finnish | Finer-data | P, L, O, Product, Event, Date | 193k |
| Greek | Legislation Corpus | P, L, O, Geopolitical entity, Public Document, Legislation | 615,000k |
| Greek | Spacy Greek | P, L, O, Geo, Event, Product | 869k |
| Hungarian | HunNERwiki | P, L, O, Miscellaneous | 562k |
| Polish | Poleval-NER 2018 | 2-level Hierarchy of 14 tags | 1,000k |
| Portuguese | LeNER-Br | P, L, O, Time, Legislation, Jurisprudence | 317k |
| Portuguese | First Harem | P, L, O, Abstraction, Event, Thing, Works, Time, Value, Other | 520k |
| Portuguese | Sigarra | P, L, O, Date, Organic Unit, Time, Course, Event | 185k |
| Romanian | Ronec | P, L, O, Ordinal, Numeric value, Date, Time, Product, Geopolitical entity, nationalities or religious or political groups, Facility, Quantity, Money, Event, Period, Work of Art, Language | 5k sentences |
| Slovenian | Slovene News Corpus | P, L, O, Miscellaneous | 6k |
| Swedish | Swedish NER Corpus | P, O, L, Miscellaneous | 155k |

Table 7: Description of selected datasets in terms of entities tags and size.

And the following table describes the various models along with the dataset and their Precision, Recall and F1 measure for the average of the 3 classes.

| Language | Tool | Dataset | P (%) | R (%) | F1 (%) |
|---|---|---|---|---|---|
| Croatian | Polyglot | Hr500k | 74.3 | 53.4 | 62.2 |
| Croatian | Polyglot | Vjesnik | 42.3 | 69.2 | 52.5 |
| Croatian | Croatian NERC System | Hr500k | 84.0 | 51.8 | 64.0 |
| Croatian | Croatian NERC System | Vjesnik | 81.0 | 54.7 | 65.4 |
| Czech | NER BERT | Czech Named Entity Corpus 2.0 | 81.0 | 75.6 | 77.8 |
| Danish | Stanford Daner | DDT | 58.4 | 71.6 | 64.3 |
| Danish | Stanford Daner | WikiANN | 56.7 | 31.8 | 40.7 |
| Estonian | ESNLTK | NER-tagger corpus | 11.7 | 73.4 | 20.7 |
| Finnish | finnish-tagtools-1.4.0 | Finer-data | 90.9 | 86.7 | 88.7 |
| Greek | Polyglot | Spacy Greek Dataset | 28.8 | 75.3 | 41.7 |
| Greek | Polyglot | Greek Legislation Corpus | 12.4 | 10.6 | 11.4 |
| Greek | Spacy Greek | Greek Legislation Corpus | 32.3 | 29.0 | 30.5 |
| Hungarian | Polyglot | HunNERwiki | 78.0 | 35.0 | 48.3 |
| Polish | poldeepner-bigru-300 | Poleval ner 2018 | 87.1 | 77.7 | 82.2 |
| Polish | poldeepner-plain-lstm-300 | Poleval ner 2018 | 0.4 | 0.1 | 0.2 |
| Polish | poldeepner-10orth-bigru | Poleval ner 2018 | 86.7 | 79.1 | 82.7 |
| Polish | poldeepner-bigru-300-mC50 | Poleval ner 2018 | 1.1 | 3.1 | 2.0 |
| Polish | poldeepner-plain-lstm-300-mC50 | Poleval ner 2018 | 81.9 | 76.3 | 79.0 |
| Portuguese | BILSTM CRF CHAR | LeNER-Br | 90.5 | 86.3 | 88.4 |
| Portuguese | BILSTM CRF CHAR | First Harem | 59.1 | 37.9 | 46.2 |
| Portuguese | BILSTM CRF CHAR | Sigarra | 39.9 | 58.4 | 47.3 |
| Portuguese | Spacy pt_core_news_sm | LeNER-Br | 26.7 | 60.8 | 37.1 |
| Portuguese | Spacy pt_core_news_sm | First Harem | 46.9 | 58.4 | 52.0 |
| Portuguese | Spacy pt_core_news_sm | Sigarra | 32.7 | 58.4 | 41.9 |
| Romanian | Polyglot | Ronec | 1.4 | 23.0 | 19.7 |
| Slovene | Stanford Model | Slovene News Corpus | 79.8 | 52.6 | 63.2 |
| Slovene | Polyglot | Slovene News Corpus | 73.2 | 29.1 | 41.7 |
| Swedish | Stagger | Swedish NER Corpus | 82.0 | 85.4 | 83.6 |
| Swedish | Swener | Swedish NER Corpus | 89.1 | 85.4 | 87.2 |

Table 8: Performance scores of NERC selected tools.

## 5. Discussion

Compared to what has been reported in the literature, we can analyse separately each platform tested.

In the case of StanfordNLP, we have found discrepant results for 6 languages: Croatian, Czech, Finnish, Greek, Latvian and Polish. Our results are worse than the reported ones for lemmatisation, PoS/MSD-tagging and dependency parsing for Greek, Polish and Portuguese. The latter being the one with higher differences (more than 10% points for LAS, for example). For Finnish and Latvian, only two tasks showed some negative difference, PoS/MSD-tagging and dependency parsing for the first and lemmatisation and PoS/MSD-tagging for the other one. For the Czech language, our LAS and UAS values are lower than the ones from the literature for two different corpora: Fictree and PDT. Croatian is the only case for which we have a better result for both lemmatisation and dependency parsing, even though we obtained inferior values for metrics concerning PoS/MSD-tagging for this language as StanfordNLP did not annotate the text in terms of morphosyntax (the XPOS test value and therefore ALLTAGS value) was zero. For the other 9 languages, results were very similar.

For the UDPipe platform, results are reproducible in all tasks for almost all the fifteen tested languages and for all existing models and corpora. The only exceptions concern worse result for PoS/MSD-tagging for Croatian, better result in this metric and in the dependency parsing evaluation for Latvian, and some discrepancy for both FTB and TDT corpora for Finnish: worse results in terms of lemmatisation metric, better results in the PoS/MSD-tagging task for FTB but worse for TDT and inferior results for LAS and UAS for TDT.

As mentioned before, NLP-Cube results were not reproducible as the published results on the platform website do not consider the whole processing chain starting from raw text. Boroş et al. (2018) propose more detailed results from raw text to CoNLL-U files for NLP-Cube models, however, there is no explicit correspondence between the models described in the article and the ones available in the platform website.

In terms of processing speed, we can observe that UDPipe is the fastest platform (almost sixty-eight times faster than NLP-Cube and seven times faster than StanfordNLP). UDPipe also has the highest standard deviation values, however, in terms of percentage of the mean speed they represent less than what is observed for StanfordNLP, which is the platform with the most significant deviation and, therefore, with higher dependency on the model and test data used.

StanfordNLP has usually relatively better metrics results compared to UDPipe and NLP Cube, however, it is slower than UDPipe in terms of processing speed. When choosing a platform one must decide if the priority will be its speed or accuracy.

Croatian, Finnish and Latvian languages are the ones with discrepancies in both UDPipe and StanfordNLP platforms, further investigation on the reasons of these differences should be conducted, also, with a detailed analysis to explain the strong differences concerning Portuguese when using StanfordNLP. Training models from StanfordNLP and NLP-Cube come from different versions of UD, therefore a detailed comparative analysis about possible existing differences inside the training and test sets between these versions could provide some input on the observed discrepancies.

In the case of Croatian for the NERC track, we compare the performance of Polyglot (Al-Rfou et al. 2015) which is built using Wikipedia and Freebase. There was no F1-score reported for the Croatian language by the author using this tool. We found it to be 52% and 62% for the datasets. Two NER models have been reported by Ljubešić et al 2013 which have 89.9% for 3-class and 63.6% for the 4-class scenario. For Croatian, we also tested the system which is FST-based (Bekavac and Tadić, 2007) and the results were in the range of 64-65% for the test set. As per our experiments, this model performed better than the polyglot model for the news domain documents.

For Czech, we used the BERT-NER (Arkhipov et al. 2019) with a multilingual model which has been reported as 93.9% partial relaxed F-1 for 5 entities. But on CNEC 2.0 the value obtained is 77.8%. The tool NameTag (Straková et al. 2013) already trained in CNEC reports a value of 82.82% F1 measure. Danish was tested on Stanford based Daner using Danish Dependency Treebank (Buch-Kromann et al. 2003) and WikiANN (Pan et al. 2017). The Multilingual BERT trained by MITP[5] is reported to have a score of 80.37% for 3-class recognition. ESTNLTK (Orasmaa et al. 2016) had a high recall but low precision resulting in poor F1.

For Finnish, we tested the finnish-tagtools-1.4.0[6] which is a HFST based tool that has decent performance compared to the rest of the available tools. For Greek, polyglot performed badly both on the News documents as well as the legislation documents.

HunnerWiki on Polyglot has low recall but high precision. PolDeepNer for 3 different models was tested on the dataset provided as part of the PolEval-2018 shared task and the results matched the reported figures. The Bi-LSTM CRF model trained on Portuguese (pt) Legal documents were used for NERC on pt-news corpus. As expected it did not perform well on the test set but its performance was well within the range that authors claim on legal documents. But the spaCy Portuguese model trained on WikiNER did not perform well on the News genre.

The worst performing model was Polyglot on the Romanian dataset which is part of the Romanian spaCy Model. As for Slovene, a 4-class Stanford model along with Polyglot was tested on News Corpus and was found to have scores of 63% and 41% F-1 respectively.

---

[5] https://mipt.ru/english/
[6] https://github.com/hfst/hfst/tree/master/scripts/finnish-tagtools

For Swedish, we used the Swedish NERC Corpus on Stagger and Sweener. Swener is a HFST based tool and as expected it performs well on the test set.

Coming to the datasets used from various sources, not all were single standardized forms.

They are either in CONLL or inline XML `<ENAMEX TYPE="ORG">Disney</ENAMEX> is a global brand.`). Datasets like Czech Dependency Treebank which had nested hierarchy were converted to CoNLL format with the aid of a script[7]. So, the CoNLL format could be employed for representing all NERC datasets. Next, most of the time, there are no direct ways to reimplement or verify the claims about the official published results, as train, development and test sets are not provided. Lastly, every language presents a different set of tags. These tags can be standardized by employing a NERC hierarchy system (Sekine 2002, Sekine et al. 2004) which can vary from coarse to a fine level.

## 6. Conclusions and Future Directions

We have presented the evaluation campaign for fifteen EU-official under-resourced languages which was conducted for testing the existing LPCs in three different platforms (Stanford CoreNLP, NLP Cube, UDPipe). The integral tasks of LPCs for each language were tokenisation, sentence splitting, PoS/MSD-tagging, lemmatisation, NERC, dependency parsing. We presented the whole results in an online available full material and in this paper only the selected data where the reproducibility of the results was in question. The criterion for this was the discrepancy that would differ for more than one percentage point from the previously reported results.

Regarding future plans, our project obligation is to enlarge the campaign to the remaining EU-official languages with special emphasis on the under-resourced ones (e.g. Maltese or Lithuanian) and to build LPCs for under-resourced languages that would respect the state-of-the-art. However, we could also apply the same approach to some languages outside of planned 24 with possible candidates selected from the list of larger languages (e.g. Russian, Chinese, Farsi, Arabic, Swahili, etc.).

Since we have observed the difficulties with evaluation of the NERC task for different languages, mainly due to the absence of universally or cross-lingually applicable named entities classification scheme, that could serve the NERC task in different languages analogous to the Universal Dependency scheme in parsing task, we decided to build such a cross-lingually usable NERC classification scheme, as one of our future research directions, and it is presented in (Alves et al., in press).

## 7. Acknowledgements

The work presented in this paper has received funding from the European Union's Horizon 2020 research and innovation programme under the Marie Skłodowska-Curie grant agreement no. 812997 and under the name CLEOPATRA (Cross-lingual Event-centric Open Analytics Research Academy).

## 8. Bibliographical References

---

[7] https://nlp.fi.muni.cz/en/AdvancedNlpCourse/NamedEntityRecognition

### 9. Language Resource References